\documentclass[11pt,a4paper]{article}
\usepackage[hyperref]{emnlp2018_arxiv}
\usepackage{times}
\usepackage{latexsym}

\usepackage{times}
\usepackage{helvet}
\usepackage{courier}
\usepackage{amsmath, amssymb}
\usepackage{bm}
\usepackage{float}
\usepackage{graphicx}
\usepackage{caption}
\usepackage{subfig}
\usepackage{multirow}
\usepackage{algorithm}
\usepackage[noend]{algpseudocode}
\usepackage{enumitem}
\usepackage{fancyvrb}

\aclfinalcopy

\title{Discrete Structural Planning for Neural Machine Translation}

\author{Raphael Shu \\
The University of Tokyo \\
\texttt{shu@nlab.ci.i.u-tokyo.ac.jp}
\And Hideki Nakayama \\
The University of Tokyo \\
\texttt{nakayama@ci.i.u-tokyo.ac.jp}
}

\date{\today}

\begin{document}
\maketitle
\begin{abstract}
Structural planning is important for producing long sentences, which is a missing part in current language generation models. In this work, we add a planning phase in neural machine translation to control the coarse structure of output sentences. The model first generates some planner codes, then predicts real output words conditioned on them. The codes are learned to capture the coarse structure of the target sentence. In order to obtain the codes, we design an end-to-end neural network with a discretization bottleneck, which predicts the simplified part-of-speech tags of target sentences. Experiments show that the translation performance are generally improved by planning ahead. We also find that translations with different structures can be obtained by manipulating the planner codes.
\end{abstract}

\section{Introduction}

When human speaks, it is difficult to ensure the grammatical or logical correctness without any form of planning. Linguists have found evidence through speech errors or particular behaviors that indicate speakers are planning ahead \cite{redford2015handbook}. Such planning can happen in discourse or sentence level, and sometimes we may notice it through inner speech.

In contrast to human, a neural machine translation (NMT) model does not have the planning phase when it is asked to generate a sentence. Although we can argue that the planning is done in the hidden layers, however, such structural information remains uncertain in the continuous vectors until the concrete words are sampled. In tasks such as machine translation, a source sentence can have multiple valid translations with different syntactic structures. As a consequence, in each step of generation, the model is unaware of the ``big picture'' of the sentence to produce, resulting in uncertainty of the choice of words.

\begin{figure}[t]
  \includegraphics[width=0.5\textwidth]{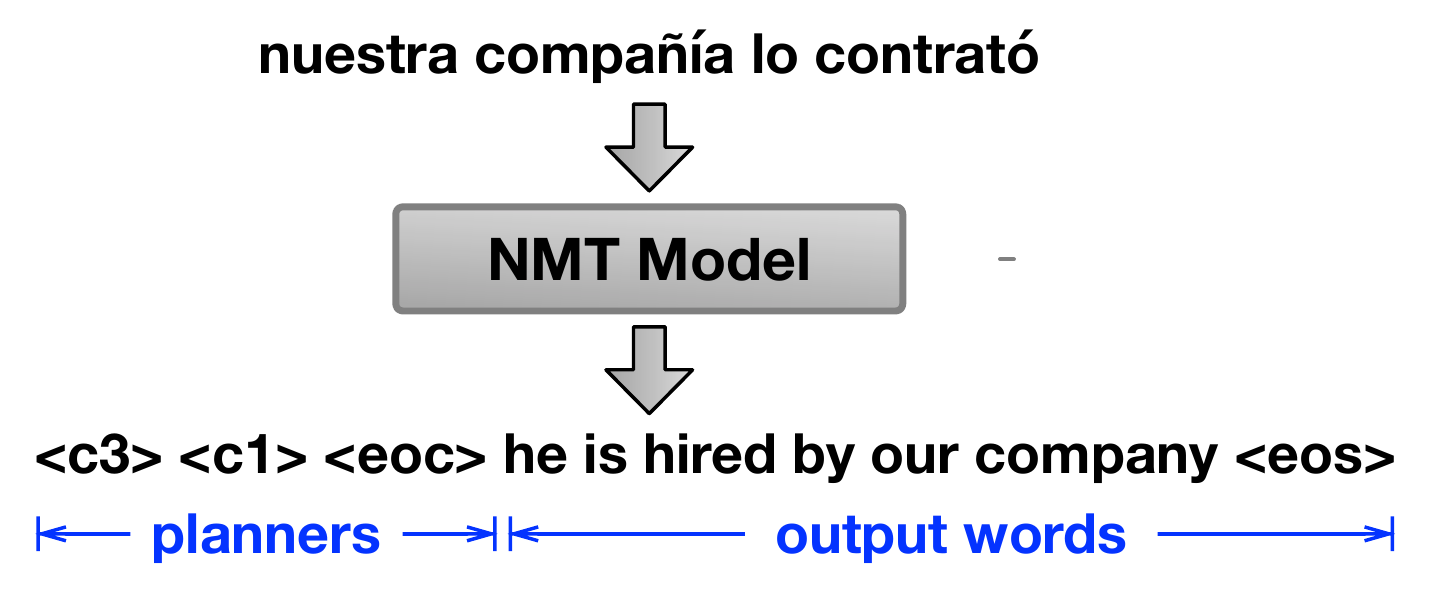}
  \caption{Illustration of the proposed sentence generation framework. The model predicts the planner codes before generating real output words.}
  \label{fig:illustrate}
\end{figure}

In this research, we try to let the model plan the coarse structure of the output sentence before decoding real words. As illustrated in Fig.~\ref{fig:illustrate}, in our proposed framework, we insert some planner codes into the beginning of the target sentences. The sentence structure of the translation is governed by the codes.

 
An NMT model takes an input sentence $X$ and produce a translation $Y$. Let $S_Y$ denotes the syntactic structure of the translation.
Indeed, the input sentence already provides rich information about the target-side structure $S_Y$. 

For example, given the Spanish sentence in Fig.~\ref{fig:illustrate}, we can easily know that the translation will have a noun, a pronoun and a verb. Such obvious structural information does not have uncertainty, and thus does not require planning. In this example, the uncertain part  is the order of the noun and the pronoun.  Thus, we want to learn a set of planner codes $C_Y$ to disambiguate such uncertain information about the sentence structure. By conditioning on the codes, we can potentially improve the effectiveness of beam search as the search space can be properly regulated.

In this work, we use simplified POS tags to annotate the structure $S_Y$. We learn the planner codes by putting a discretization bottleneck in an end-to-end network that reconstructs $S_Y$ with both $X$ and $C_Y$. The codes are merged with the target sentences in the training data. Thus, no modification to the NMT model is required. Experiments show the translation performance is generally improved with structural planning. More interestingly, we can control the structure of output sentences by manipulating the planner codes.

\section{Learning Structural Planners}

In this section, we first extract the structural annotation $S_Y$ by simplifying the POS tags. Then we explain the code learning model for obtaining the planner codes.

\subsection{Structural Annotation with POS Tags}

To reduce uncertainty in the decoding phase, we want a structural annotation that describes the ``big picture'' of the sentence. For instance, the annotation can tell whether the sentence to generate is in a ``NP VP'' order. The uncertainty of local structures can be efficiently solved by beam search or the NMT model itself.

In this work, we extract such coarse structural annotations $S_Y$ through a simple two-step process that simplifies the POS tags of the target sentence:
\begin{enumerate}
    \item Remove all tags other than  ``{\bf N}'', ``{\bf V}'', ``{\bf PRP}'', ``{\bf ,}'' and ``{\bf .}''. Note that all tags begin with ``N'' (e.g.~{\bf NNS}) are mapped to ``{\bf N}'', and tags begin with ``V'' (e.g.~{\bf VBD}) are mapped to ``{\bf V}''.
    \item Remove duplicated consecutive tags.
\end{enumerate}
The following list gives an example of the process:
\begin{small}
\begin{Verbatim}[commandchars=\\\{\},]
   Input: \textbf{He found a fox behind the wall.}
POS Tags: \textbf{PRP VBD DT NN IN DT NN .}
  Step 1: \textbf{PRP V N N .}
  Step 2: \textbf{PRP V N }.
\end{Verbatim}
\end{small}
Note that many other annotations can also be considered to represent the syntactic structure, which is left for future work to explore. 

\subsection{Code Learning}

\begin{figure}[t]
  \includegraphics[width=0.5\textwidth]{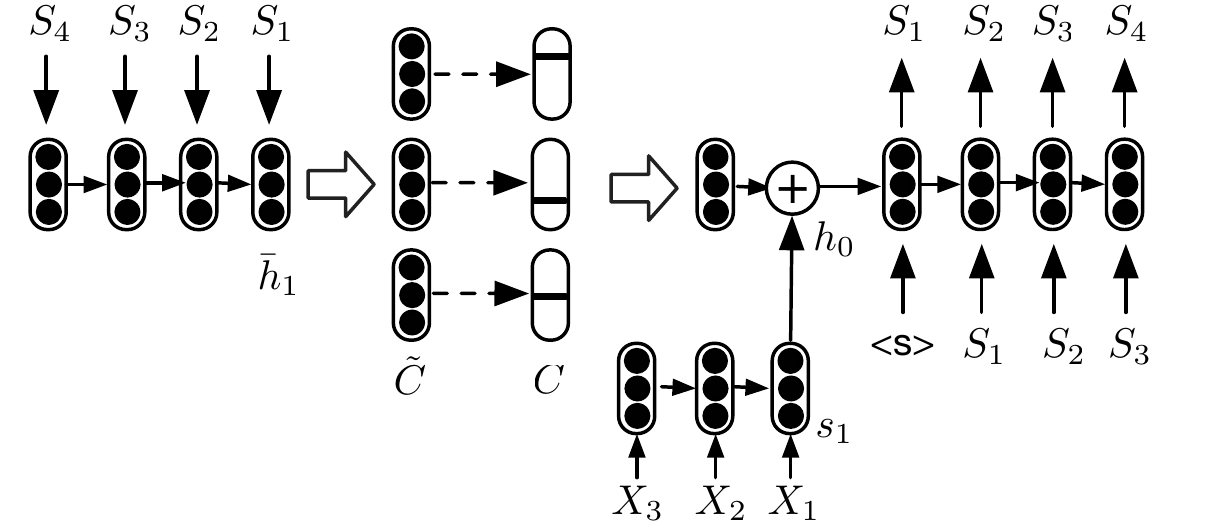}
  \caption{Architecture of the code learning model. The discretization bottleneck is shown as the dashed lines.}
  \label{fig:learn}
\end{figure}

Next, we learn the planner codes $C_Y$ to remove the uncertainty of the sentence structure $S_Y$ when producing a translation.
For simplicity, we use the notion $S$ and $C$ to replace $S_Y$ and $C_Y$ in this section.

We first compute the discrete codes $C_1,..,C_N$ based on simplified POS tags $S_1,...,S_T$:
\begin{align}
    \bar h_{\mathrm{t}} &= \mathrm{LSTM}(\mathrm{E}(S_t), \bar h_{t+1}; \theta_{\mathrm{s}}) \:, \\
    [\tilde C_1,...,\tilde C_N] &= f_{\mathrm{enc}}(\bar h_1;\theta_{\mathrm{enc}}) \:, \\
    C_i &= \mathrm{GumbelSoftmax}(\tilde C_i) \:,
\end{align}
where the tag sequence $S_1,...,S_T$ is firstly encoded using a {\it backward} LSTM \cite{hochreiter1997long}. $\mathrm{E}(\cdot)$ denotes the embedding function. Then, we compute a set of vectors $\tilde C_1, ..., \tilde C_N$, which are latterly discretized in to approximated one-hot vectors $C_1, ..., C_N$ using Gumbel-Softmax trick \cite{Jang2016CategoricalRW,Maddison2016TheCD}.

We then combine the information from $X$ and $C$ to initialize a decoder LSTM that sequentially predicts $S_1,..., S_T$:
\begin{align}
    s_t &= \mathrm{LSTM}(\mathrm{E}(X_t), s_{t+1}; \theta_{x}) \:, \\
    h_0 &= f_{\mathrm{dec}}([C_1,...,C_N];\theta_{\mathrm{dec}}) + s_1 \:, \\
    h_t &= \mathrm{LSTM}(\mathrm{E}(S_{t-1}), h_{t-1}; \theta_{h}) \:,
\end{align}
where $[C_1,...,C_N]$ denotes a concatenation of $N$ one-hot vectors. Note that only $h_t$ is computed with a {\it forward} LSTM. Both $f_{\mathrm{enc}}$ and $f_{\mathrm{dec}}$ are affine transformations. Finally, we predict the probability of emitting each tag $S_t$ with
\begin{align}
    P(&S_t|S_{1:t-1}, X, C) = \mathrm{softmax}(f_{\mathrm{out}}(h_t; \theta_{\mathrm{out}})) \:.
\end{align}

The architecture of the code learning model is depicted in Fig.~\ref{fig:learn}, which can be seen as a sequence auto-encoder with an extra context input $X$ to the decoder. The parameters are optimized with cross-entropy loss.

Once the code learning model is trained, we can obtain the planner codes $C$ for all target sentences in the training data using the encoder part.



\section{NMT with Structural Planning}

The training data of machine translation dataset is composed of $(X, Y)$ sentence pairs. With the planner codes $C_Y$ we obtained, our training data now becomes a list of $(X, C_Y;Y)$ pairs. As shown in Fig.~\ref{fig:illustrate}, we connect the planner codes and target sentence with a ``$\langle\text{eoc}\rangle$'' token.

With the modified dataset, we train a regular NMT model. We use beam search when decoding sentences, thus the planner codes are searched before emitting real words. The codes are removed from the translation results during evaluation.

\section{Related Work}

Recently, some methods are proposed to improve the syntactic correctness of the translations. \citet{Stahlberg2016SyntacticallyGN} restricts the search space of the NMT decoder using the lattice produced by a Statistical Machine Translation system. \citet{Eriguchi2017LearningTP} takes a multi-task approach, letting the NMT model to parse a dependency tree and combine the parsing loss with the original loss.

Several works further incorporate the target-side syntactic structures explicitly. \citet{Nadejde2017PredictingTL} interleaves CCG supertags with normal output words in the target side. Instead of predicting words, \citet{Aharoni2017TowardsSN} trains a NMT model to generate linearized constituent parse trees. \citet{Wu2017SequencetoDependencyNM} proposed a model to generate words and parse actions simultaneously. The word prediction and action prediction are conditioned on each other. However, none of the these methods plan the structure before translation.

Similar to our code learning approach, some works also learn the discrete codes for different purposes. \citet{Shu2017CompressingWE} compresses the word embeddings by learning the concept codes to represent each word. \citet{Kaiser2018FastDI} breaks down the dependency among words with shorter code sequences. The decoding can be faster by predicting the shorter artificial codes.

\section{Experiments}

We evaluate our models on IWSLT 2014 German-to-English task \citep{cettolo2014report} and ASPEC Japanese-to-English task \citep{NAKAZAWA16.621}, containing 178K and 3M bilingual pairs respectively. We use Kytea \citep{kytea} to tokenize Japanese texts and moses toolkit \citep{Koehn2007MosesOS} for other languages. Using byte-pair encoding \citep{Sennrich2016NeuralMT}, we force the vocabulary size of each language to be 20K for IWSLT dataset and 40K for ASPEC dataset. 

For IWSLT 2014 dataset, we concatenate all five TED/TEDx development and test corpus to form a test set containing 6750 pairs.  For evaluation, we report {\it tokenized BLEU} with moses tool.

\subsection{Evaluation of Planner Codes}

In the code learning model, all hidden layers have 256 hidden units. The model is trained using Nesterov's accelerated gradient (NAG) \citep{nesterov1983method} for maximum 50 epochs with a learning rate of $0.25$. We test different settings of code length $N$ and the number of code types $K$. The information capacity of the codes will be $N \log K$ bits. In Table \ref{table:code}, we evaluate the learned codes for different settings. $S_y$ accuracy evaluates the accuracy of correctly reconstructing $S_y$ with the source sentence $X$ and the code $C_y$. $C_y$ accuracy reflects the chance of guessing the correct code $C_y$ given $X$.

\begin{table}[t]
\begin{center}
    \begin{tabular}{c|c|c|c}
    \hline \hline
    {\bf Code Setting} & {\bf Capacity} & {\bf $S_Y$ acc.} & {\bf $C_Y$ acc.} \\
    \hline
    N=1, K=4 & 2 bits & 27\% & 63\% \\
    N=2, K=2 & 2 bits & 23\% & 67\% \\
    N=2, K=4 & 4 bits & 35\% & 41\% \\
    N=4, K=2 & 4 bits & 22\% & 44\% \\
    N=4, K=4 & 8 bits & 44\% & 27\% \\
    \hline \hline
    \end{tabular}
    \caption{A comparison of different code settings on IWSLT 2014 dataset. The accuracy of reconstructing $S_Y$ in the code model, and the accuracy of predict $C_Y$ in the NMT model are reported.}
    \label{table:code}
\end{center}
\end{table}

We can see a clear trade-off between $S_Y$ accuracy and $C_Y$ accuracy. When the code has more capacity, it can recover $S_Y$ more accurately, however, resulting in a lower probability for the NMT model to guess the correct code. We found the setting of $N=2, K=4$ has a balanced  trade-off.

\subsection{Evaluation of NMT Models}

To make a strong baseline, we use 2 layers of bi-directional LSTM encoders with 2 layers of LSTM decoders in the NMT model. The hidden layers have 256 units for IWSLT De-En task and 1000 units for ASPEC Ja-En task. We apply Key-Value Attention \citep{Miller2016KeyValueMN} in the first decoder layer. Residual connection \citep{He2016DeepRL} is used to combine the hidden states in two decoder layers. Dropout is applied everywhere outside of the recurrent function with a drop rate of $0.2$ . To train the NMT models, we also use the NAG optimizer with a learning rate of 0.25, which is annealed by a factor of 10 if no improvement of loss value is observed in 20K iterations. Best parameters are chosen on a validation set.

\begin{table}[t]
\begin{center}
\begin{small}
    \begin{tabular}{c|c|c|c|c}
    \hline \hline
    \multirow{2}{*}{\bf Dataset} & \multirow{2}{*}{\bf Model} & \multicolumn{3}{c}{{\bf BLEU(\%)}} \\
    \cline{3-5}
     & & {\bf BS=1} & {\bf BS=3} & {\bf BS=5} \\
    \hline
    \multirow{2}{*}{De-En} & baseline & 27.90 & 29.26 & 29.52 \\
    & {\small plan (N=2, K=4)} & {\bf 28.35} & {\bf 29.59} & {\bf 29.78} \\
    \hline
    \multirow{2}{*}{Ja-En} & baseline & {\bf 23.92} & 25.08 & 25.26 \\
    & {\small plan (N=2, K=4)} & 22.79 & {\bf 25.53} & {\bf 25.69} \\
    \hline \hline
    \end{tabular}
    \caption{A comparison of translation performance with different beam sizes (BS).}
    \label{table:translate}
\end{small}
\end{center}
\end{table}

As shown in Table~\ref{table:translate}, by conditioning the word prediction on the generated planner codes, the translation performance is generally improved over a strong baseline. The improvement may be the result of properly regulating the search space.

However, when we apply greedy search on Ja-En dataset, the BLEU score is much lower compared to the baseline. We also tried to beam search the planner codes then switch to greedy search, but the results are not significantly changed. We hypothesize that it is important to simultaneously explore multiple candidates with drastically different structures in Ja-En task. By planning ahead, more diverse candidates can be explored, which improves beam search but not greedy search. If so, the results are in line with a recent study \cite{Li2016ASF} that shows the performance of beam search depends on the diversity of candidates.

\subsection{Qualitative Analysis}

Instead of letting the beam search to decide the planner codes, we can also choose the codes manually. Table~\ref{table:manipulate} gives an example of the candidate translations produced by the model when conditioning on different planner codes.

\begin{table}[h]
\begin{center}
\begin{small}
    \begin{tabular}{c|p{6.2cm}}
    \textbf{input} & AP no katei ni tsuite nobeta. (Japanese) \\
    \hline
    \multirow{2}{*}{\textbf{code 1}} & \texttt{\textbf{<c4> <c1> <eoc>}} \\
    & \texttt{\footnotesize the process of AP is described .} \\
    \hline
    \multirow{2}{*}{\textbf{code 2}} & \texttt{\textbf{<c1> <c1> <eoc>}} \\
    & \texttt{\footnotesize this paper describes the process of AP .} \\
    \hline
    \multirow{2}{*}{\textbf{code 3}} & \texttt{\textbf{<c3> <c1> <eoc>}} \\
    & \texttt{\footnotesize here was described on process of AP .} \\
    \hline
    \multirow{2}{*}{\textbf{code 4}} & \texttt{\textbf{<c2> <c1> <eoc>}} \\
    & \texttt{\footnotesize they described the process of AP .} \\
    \end{tabular}
    \caption{Example of translation results conditioned on different planner codes in Ja-En task}
    \label{table:manipulate}
\end{small}
\end{center}
\end{table}

As shown in Table~\ref{table:manipulate}, we can obtain translations with drastically different structures by manipulating the codes. The results show that the proposed method can be useful for sampling paraphrased translations with high diversity.

\begin{figure}[t]
  \vspace{-6pt}
  \includegraphics[width=0.5\textwidth]{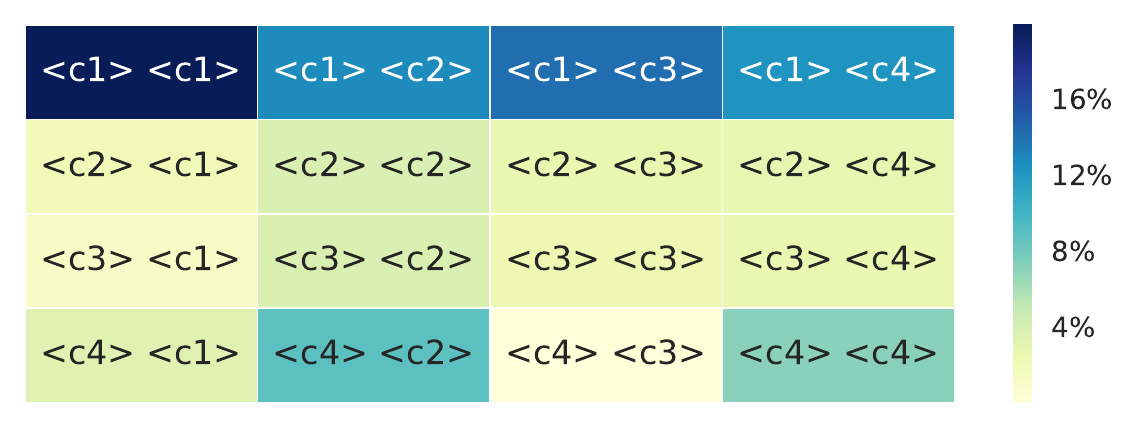}
  \caption{Distribution of assigned planner codes for English sentences in ASPEC Ja-En dataset}
  \label{fig:balance}
\end{figure}

The distribution of the codes learned for 3M English sentences in ASPEC Ja-En dataset is shown in Fig.~\ref{fig:balance}. We found the code ``\texttt{<c1> <c1>}'' is assigned to 20\% of the sentences, whereas ``\texttt{<c4> <c3>}'' is not assigned to any sentence. The skewed distribution may indicate that the capacity of the codes is not fully exploited, and thus leaves room for further improvement.

\section{Discussion}

Instead of learning discrete codes, we can also directly predict the structural annotations (e.g.~POS tags), then translate based on the predicted structure. However, as the simplified POS tags are also long sequences, the error of predicting the tags will be propagated to word generation. In our experiments, doing so degrades the performance by around 8 BLEU points on IWSLT dataset.

\section{Conclusion}

In this paper, we add a planning phase in neural machine translation, which generates some planner codes to control the structure of the output sentence. 
To learn the codes, we design an end-to-end neural network with a discretization bottleneck to predict the simplified POS tags of target sentences. Experiments show that the proposed method generally improves the translation performance. We also confirm the effect of the planner codes, by being able to sample translations with drastically different structures using different planner codes.

The planning phase helps the decoding algorithm by removing the uncertainty of the sentence structure. 
The framework described in this paper can be extended to plan other latent factors, such as the sentiment or topic of the sentence.

\bibliography{mybib}
\bibliographystyle{acl_natbib_nourl}

\newpage
\appendix
\onecolumn
\section{Examples of Generated Translations}

We show some random translation examples in ASPEC Ja-En task. The length of input sentence is limited below 10 words. The second code tends to be ``$\langle$c1$\rangle$'' because it may learns to capture information for long sentences.

\begin{small}
\begin{Verbatim}[commandchars=\\\{\},]
\textbf{Input: saigo ni， shorai tenbo ni tsu ite kijutsu .}
<c3> <c2> <eoc> finally , the future prospects are described .
<c4> <c1> <eoc> future prospects are also described .


\textbf{Input: DNA kaiseki no gijutsu wo kaisetu shi ta .}
<c4> <c1> <eoc> the technology of DNA analysis is explained .
<c1> <c1> <eoc> this paper explains the technology of DNA analysis .


\textbf{Input: ekisho tokusei hyouka souchi no shoukai de a ru .}
<c3> <c1> <eoc> this is an introduction to liquid crystal property evaluation equipment .
<c4> <c1> <eoc> the liquid crystal property evaluation equipment is introduced .

\textbf{Input: gaiyou zai de chiryou shi ta .}
<c2> <c1> <eoc> it was treated with external preparation .
<c1> <c1> <eoc> the patient was treated with external preparation .

\textbf{Input: ukeire no kahi ha shichou ga handan suru .}
<c1> <c1> <eoc> the city length is judged the propriety of the acceptance .
<c4> <c1> <eoc> the propriety of the acceptance judges the city length .

\textbf{Input: fuku sayou ha na ka ta .}
<c3> <c1> <eoc> there was no side effect .
<c4> <c1> <eoc> no side effect was observed .

\textbf{Input: heriumu reidou shisutemu no shiyo wo kaisetsu shi ta .}
<c4> <c1> <eoc> the specification of the helium refrigeration system was explained .
<c1> <c1> <eoc> this paper explains the specification of the helium refrigeration system .

\textbf{Input: kenkyu han no gaiyou wo shoukai shi ta .}
<c4> <c1> <eoc> the outline of the research team is introduced .
<c1> <c1> <eoc> this paper introduces the outline of the research team .

\textbf{Input: kahen bunsan hoshou ki wo kaihatsu shi ta .}
<c4> <c1> <eoc> a variable dispersion compensator has been developed .
<c2> <c1> <eoc> we have developed a variable dispersion compensator .


\textbf{Input: kairo no sekai to tokusei wo setumei shi ta .}
<c1> <c1> <eoc> this paper explains the design and characteristics of the circuit .
<c4> <c1> <eoc> the design and characteristics of the circuit are explained .	

\end{Verbatim}
\end{small}

\end{document}